\documentclass[9pt,twocolumn,twoside]{opticajnl}
\journal{opticajournal} 

\setboolean{shortarticle}{true}

\usepackage{amsmath,amsfonts,bm}

\def\eqref#1{equation~\ref{#1}}

\def\1{\bm{1}}

\def\vr{{\bm{r}}}

\DeclareMathAlphabet{\mathsfit}{\encodingdefault}{\sfdefault}{m}{sl}
\SetMathAlphabet{\mathsfit}{bold}{\encodingdefault}{\sfdefault}{bx}{n}

\def\gF{{\mathcal{F}}}

\def\gH{{\mathcal{H}}}

\usepackage{subfig}
\usepackage{textcomp}
\usepackage[capitalize,noabbrev]{cleveref}
\usepackage{epsfig}

\usepackage{lineno}

\title{Machine-Learning-based Colorectal Tissue Classification via Acoustic Resolution Photoacoustic Microscopy}

\author[1,4]{Shangqing Tong}
\author[1,4]{Peng Ge}
\author[2,4]{Yanan Jiao}
\author[2]{Zhaofu Ma}
\author[1]{Ziye Li}
\author[3]{Longhai Liu}
\author[1]{Feng Gao}
\author[2,$\ddagger$]{Xiaohui Du}
\author[1,$\dagger$]{Fei Gao}

\affil[1]{School of Information Science and Technology, ShanghaiTech University, 393 Middle Huaxia Road, Shanghai, 201210}
\affil[2]{The First Medical Centre, PLA General Hospital, 28 Fuxing Road, Haidian District, Beijing, 100091}
\affil[3]{New Concept Product Initiative Department, Advantest (China) Co., Ltd., Shanghai, China}
\affil[4]{The authors contributed equally to this work.}

\affil[$\dagger$]{gaofei@shanghaitech.edu.cn}
\affil[$\ddagger$]{duxiaohui301@sina.com}

\begin{abstract}
    Colorectal cancer is a deadly disease that has become increasingly prevalent in recent years. Early detection is crucial for saving lives, but traditional diagnostic methods such as colonoscopy and biopsy have limitations. Colonoscopy cannot provide detailed information within the tissues affected by cancer, while biopsy involves tissue removal, which can be painful and invasive. In order to improve diagnostic efficiency and reduce patient suffering, we studied machine-learning-based approach for colorectal tissue classification that uses acoustic resolution photoacoustic microscopy (AR-PAM). With this tool, we were able to classify benign and malignant tissue using multiple machine learning methods. Our results were analyzed both quantitatively and qualitatively to evaluate the effectiveness of our approach.
\end{abstract}

\setboolean{displaycopyright}{false} 

\begin{document}

\maketitle

Colorectal cancer (CRC) has become the second most common cancer diagnosed in the United States each year~\cite{Siegel2020ColorectalCS}. According to the report of the National Cancer Institute, it was estimated that 151,030 new cases of CRC were diagnosed in 2022, which was the cause of 52,580 deaths. The American Cancer Society reports that detecting and removing adenomatous polyps can largely prevent CRC, and survival rates are significantly better when it's diagnosed at a localized stage~\cite{Levin2008ScreeningAS}. The 5-year relative survival rate is 90\% for patients whose CRC is diagnosed at a localized stage, but drops to 14\% for those diagnosed at a distant stage.

Colonoscopy is a commonly used approach for early colorectal cancer (CRC) detection. However, it has limitations in that it can only image the surface of the colon and rectum, and cannot determine the depth of cancer tissue invasion.
During a colonoscopy, a long, flexible tube is inserted into the patient's rectum to provide real-time video of the inside tissues. However, while colonoscopy can suggest the presence of cancerous tissues, a biopsy is necessary for a definitive diagnosis. During a biopsy, a small piece of colon or rectum tissue is removed and observed under a microscope.
To accurately diagnose CRC, a biomarker test is often required to identify unique genes, proteins, and other factors specific to the tumor. Unfortunately, this type of pathology analysis requires tissue samples, which can be invasive and painful for patients during the excision. While existing clinical imaging modalities such as CT, PET/CT, ultrasound, and MRI have advantages, they still cannot accurately determine the area of cancer located deep within the colorectal tissue.

\begin{figure}[t]
    \centering
    \includegraphics[width=\columnwidth,trim={230 150 230 150},clip]{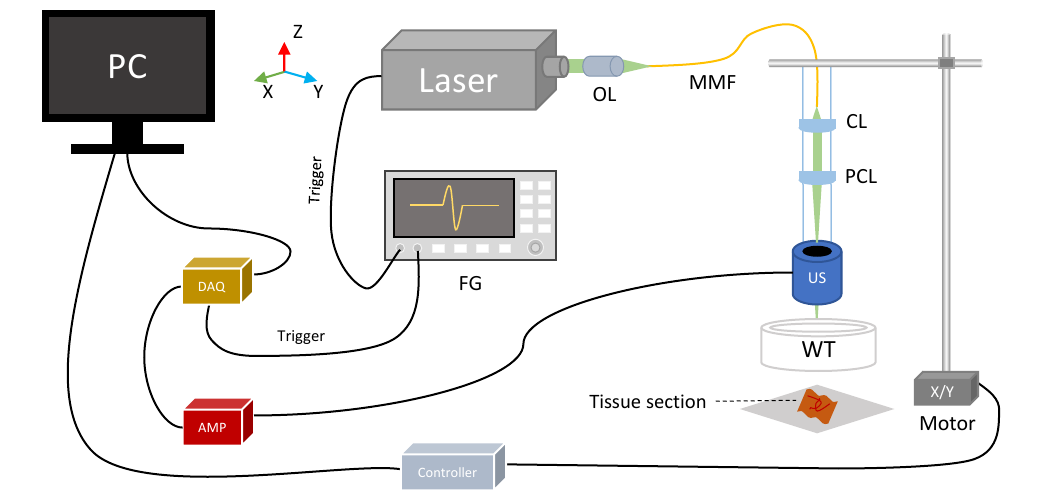}
    \caption{System setup of the Acoustic Resolution Photoacoustic Microscopy system. PC, Personal Computer; FG, Function Generator; OL, Objective Lens; MMF, Multimode Fiber; CL, Collimating Lens; PCL, Plano Convex Lens; WT, Water Tank; AMP, Amplifier; DAQ, Data Acquisition Card.}
    \label{fig:system}
\end{figure}

\begin{figure*}[t]
    \centering
    \subfloat[]{
    \begin{minipage}[t]{0.32\textwidth}
    \centering
    \includegraphics[height=0.5\textwidth]{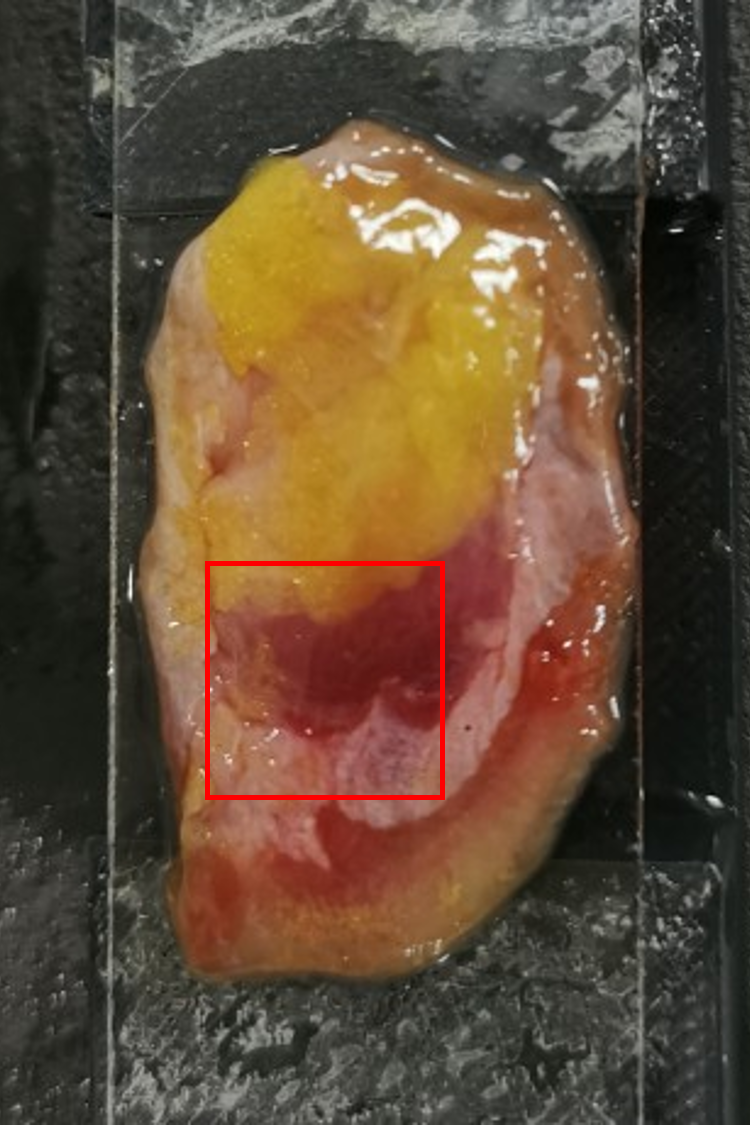}
    \includegraphics[height=0.5\textwidth]{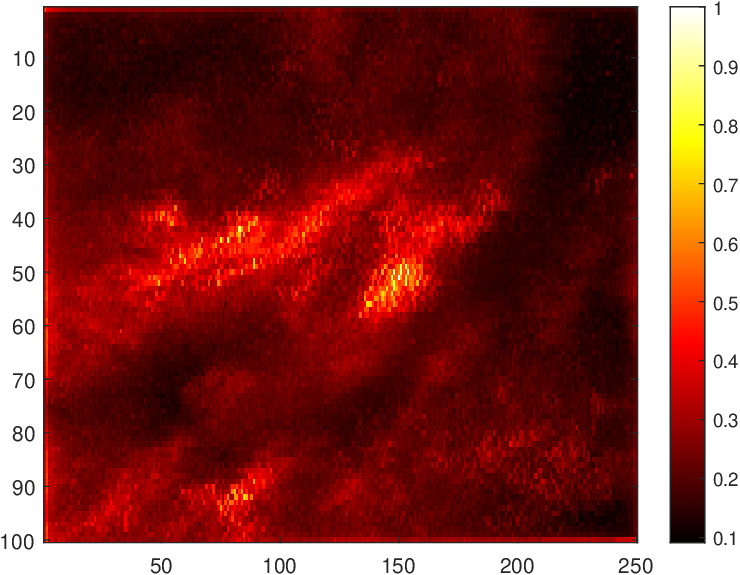} \\ \vspace{0.15mm}
    \includegraphics[height=0.5\textwidth]{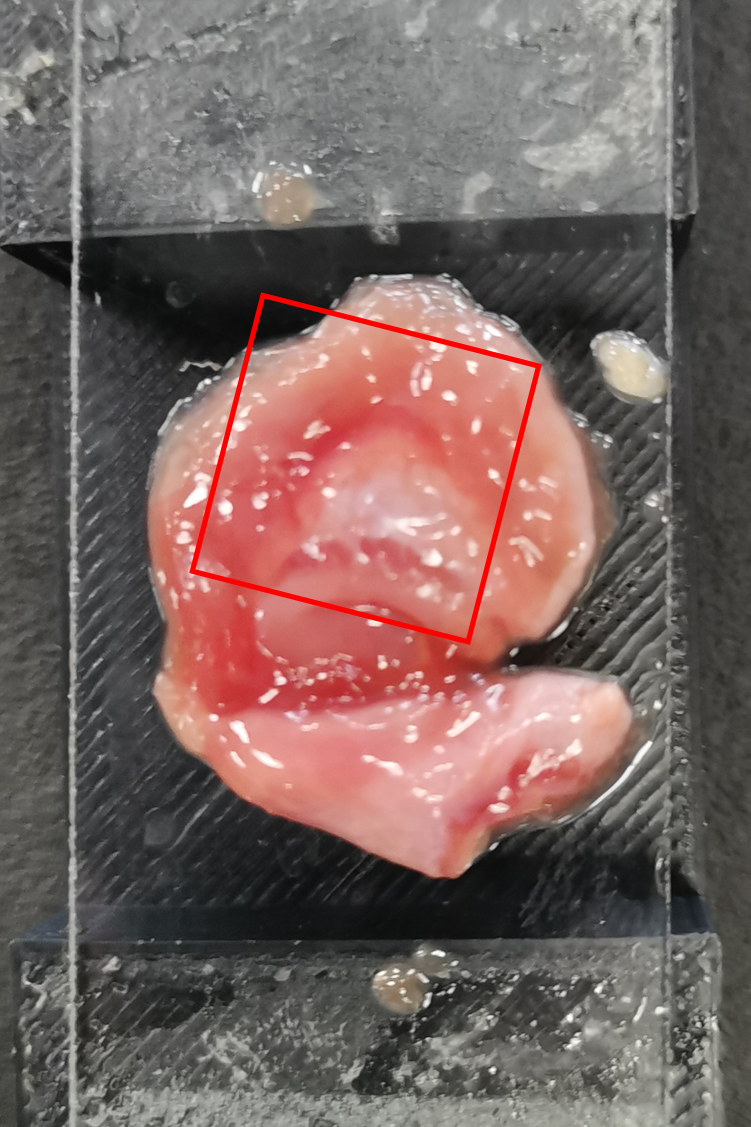}
    \includegraphics[height=0.5\textwidth]{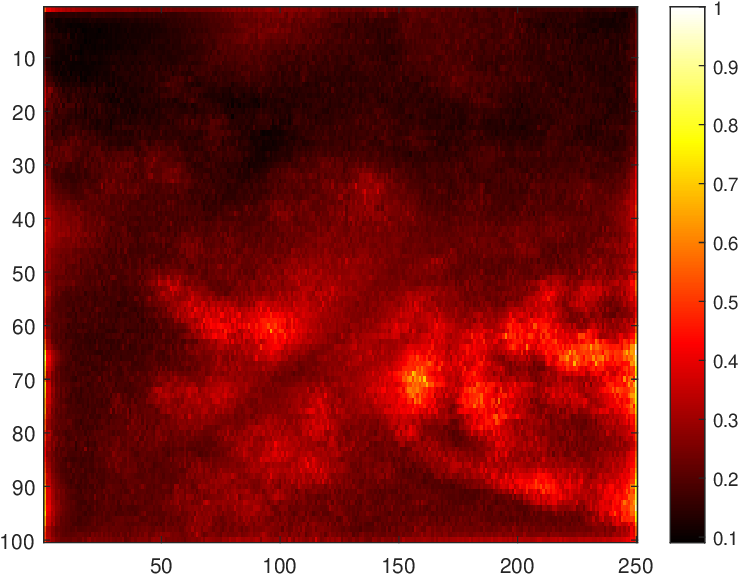}
    \end{minipage}
    }
    \subfloat[]{
    \begin{minipage}[t]{0.32\textwidth}
    \centering
    \includegraphics[height=0.5\textwidth]{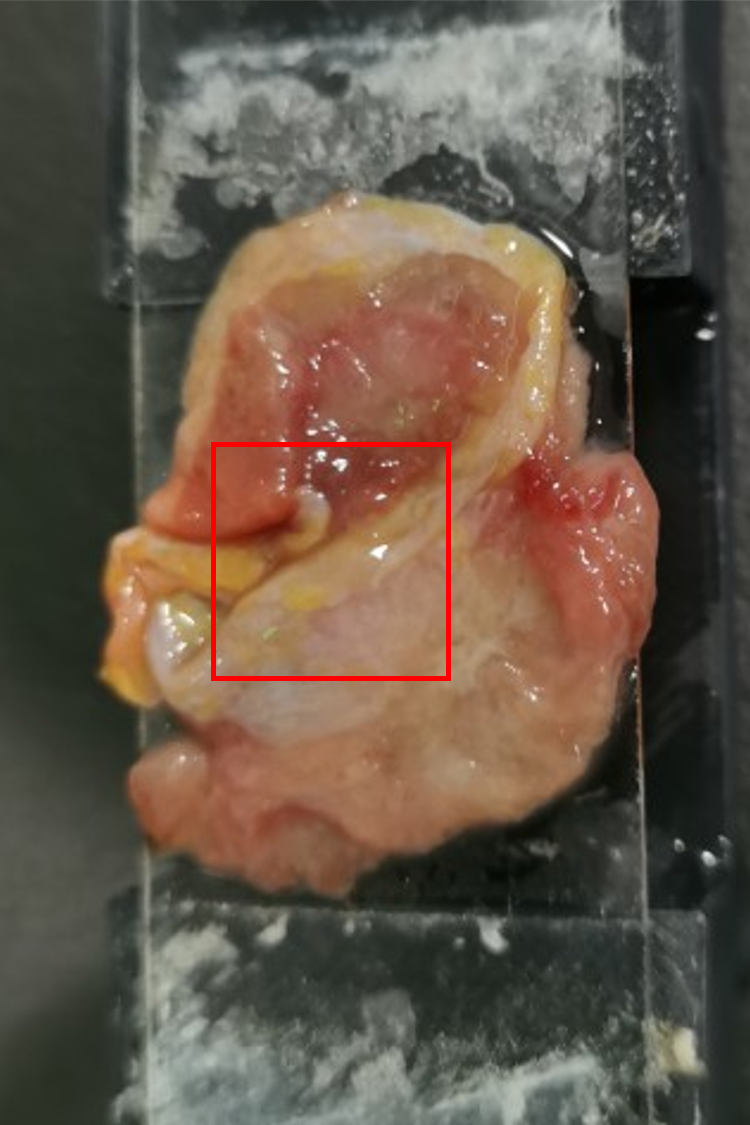}
    \includegraphics[height=0.5\textwidth]{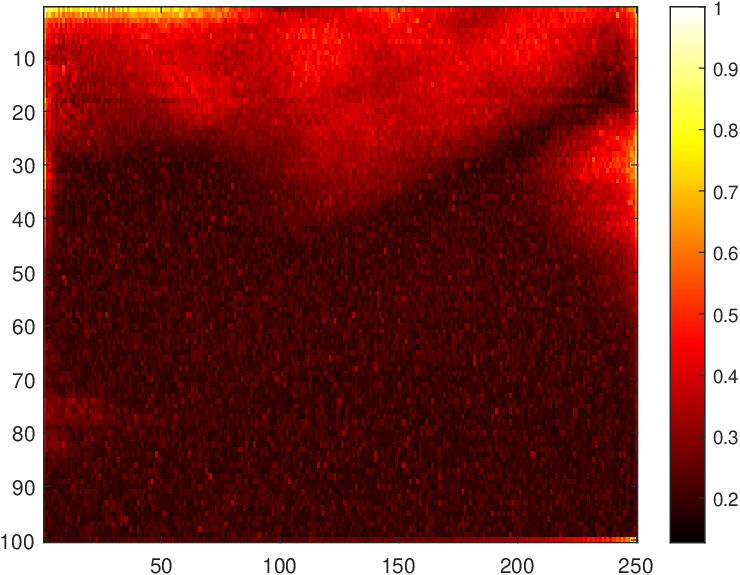}\\ \vspace{0.15mm}
    \includegraphics[height=0.5\textwidth]{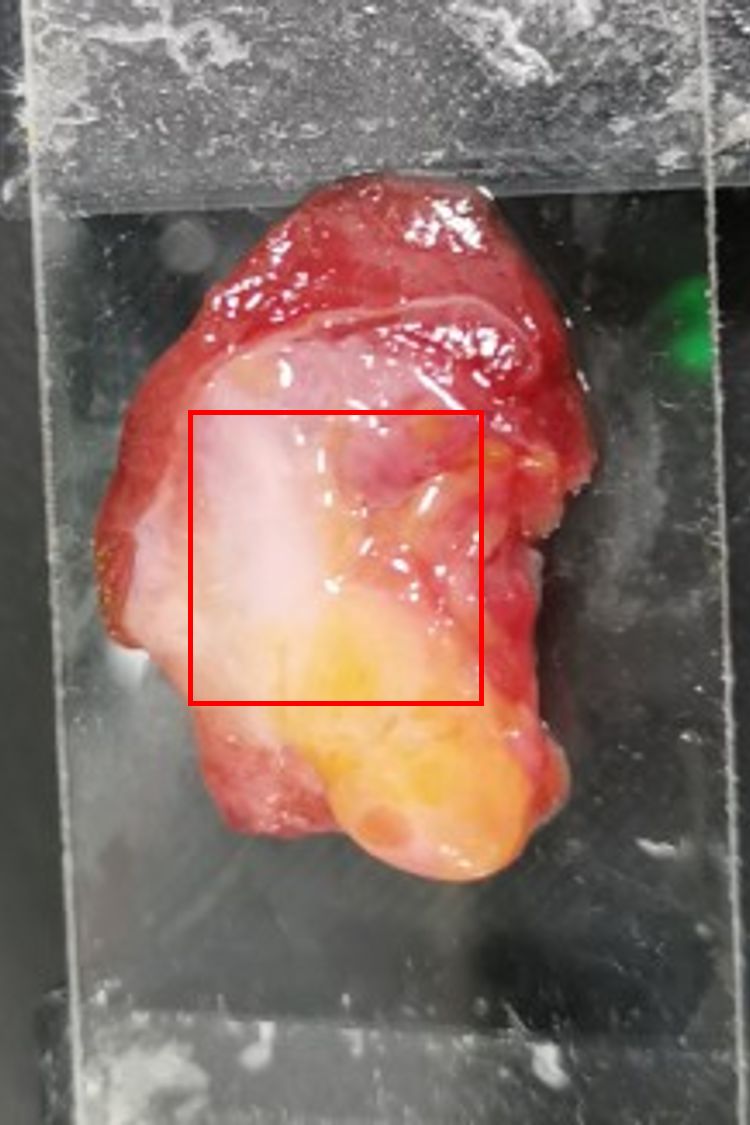}
    \includegraphics[height=0.5\textwidth]{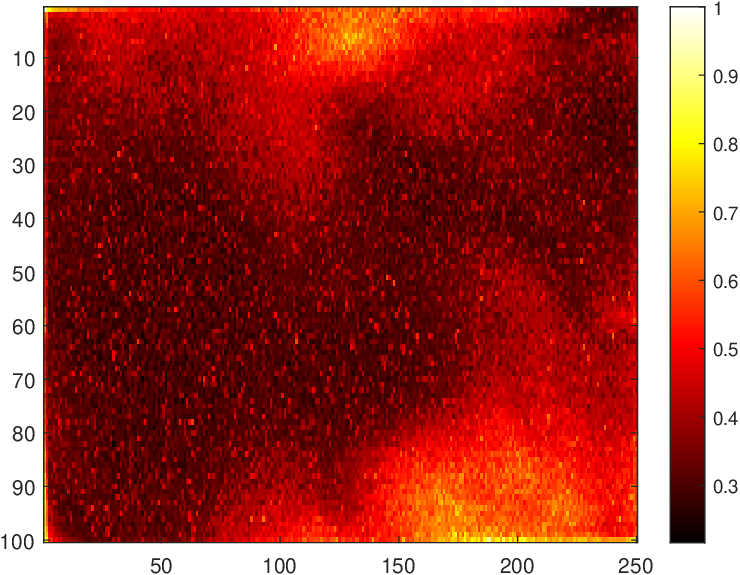}
    \end{minipage}
    }
    \subfloat[]{
    \begin{minipage}[t]{0.32\textwidth}
    \centering
    \includegraphics[height=0.5\textwidth]{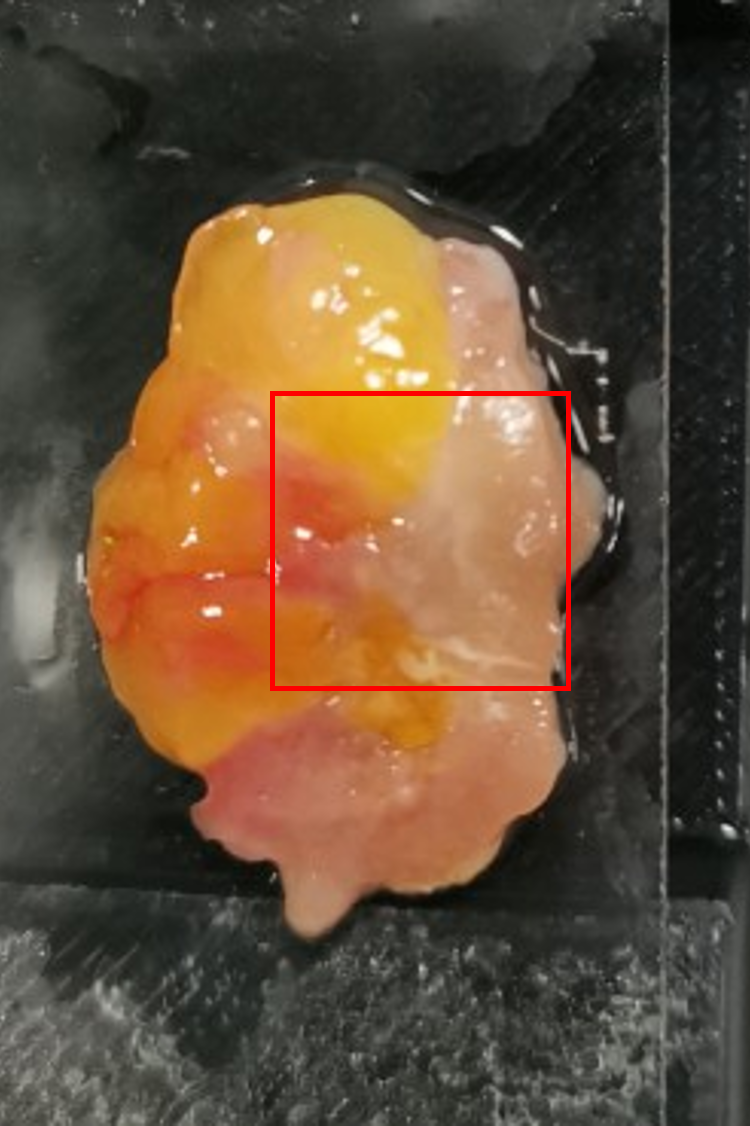}
    \includegraphics[height=0.5\textwidth]{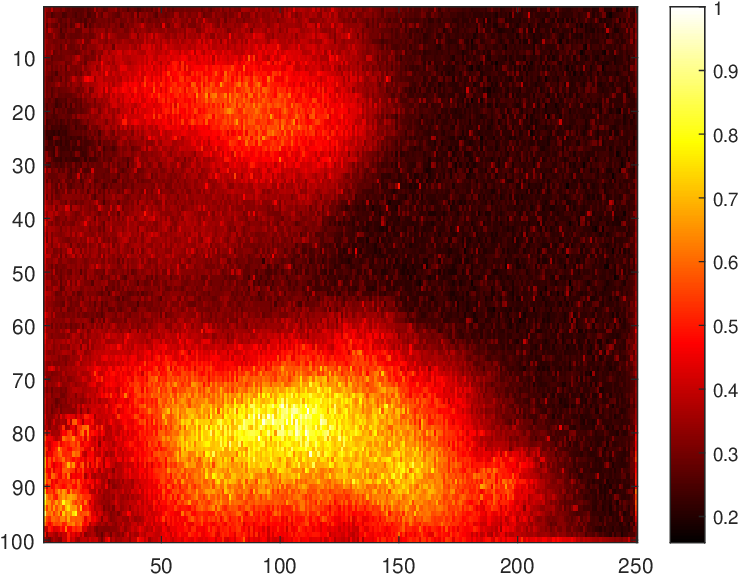}\\ \vspace{0.15mm}
    \includegraphics[height=0.5\textwidth]{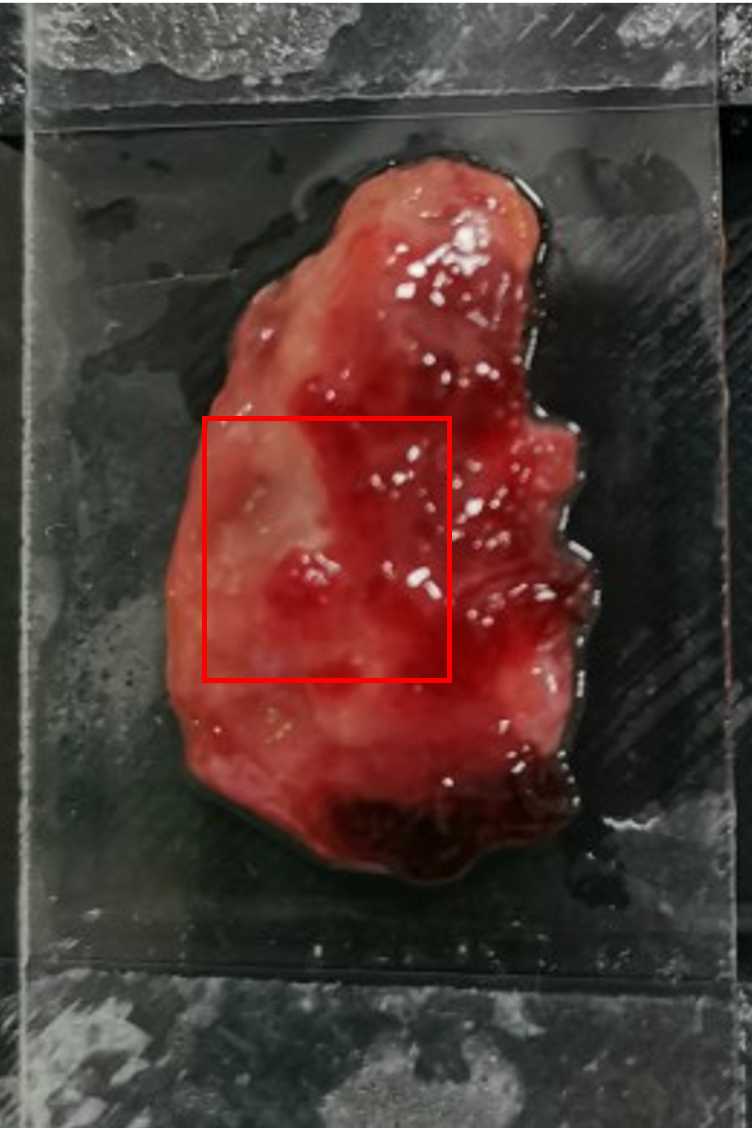}
    \includegraphics[height=0.5\textwidth]{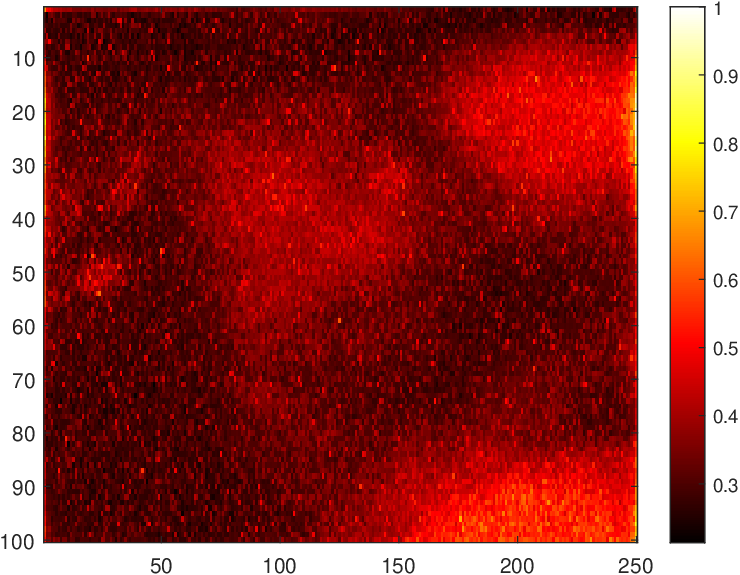}
    \end{minipage}
    }
    \caption{Photographs and imaging results for colorectal samples. \textit{(a)}, image results for normal region; \textit{(b)}, image results for border region; \textit{(c)}, image results for cancer region. The regions of interest are marked with red boxes on the photographs. Left, photograph of samples; right, maximum amplitude projection of samples.}
    \label{fig:image-results}
\end{figure*}

As a newly emerged imaging modality, photoacoustic imaging~\cite{Xu2006PhotoacousticII} combines the advantages of optical imaging and ultrasound imaging, providing high resolution and penetration depth.
Particularly, photoacoustic microscopy (PAM) is an imaging technique based on the photoacoustic effect, which makes use of the local increase in temperature that occurs as a result of the absorption of light in the tissue~\cite{Liu2018PhotoacousticMP}.
PAM screens a biological tissue by finding the local pressure increase $p$ at position $\vr$, which is the goal of photoacoustic microscopy, for it can be used to recover the absorption coefficient $\mu_a$ following
\begin{equation}
    p(\vr) = \Gamma \eta \mu_a (\vr) F(\vr),
\end{equation}
where $\eta$ is percentage of absorbed light energy converted to heat, $F(\vr)$ is the local optical fluence, $\Gamma$ is the Gr\"ueneisen parameter which is defined as
\begin{equation}
    \Gamma = \frac{\beta}{\kappa \rho C_V},
\end{equation}
where $\beta$ is the thermal coefficient of volumn expansion, $\rho$ is the density, and $\kappa$ is the isothermal compressibility. PAM has involved in various biomedical applications, including microvascular imaging, blood flow imaging and single cell imaging.

There is a wealth of literature on colorectal cancer classification using photoacoustic imaging techniques. For instance, Xiandong Leng et al. investigated the feasibility of using a dual-modality approach combining ultrasound imaging with an endoscopic AR-PAM system~\cite{Leng2018FeasibilityOC}. The PAM signals generated by the endoscopy system were evaluated using the power spectral slope derived from fast Fourier transform. Similarly, Guang Yang et al. proposed a method for imaging colorectal cancer tissues using dual-modality imaging with ultrasound and photoacoustic techniques~\cite{Yang2019CoregisteredPA}.

To take advantage of photoacoustic imaging for diagnosing colorectal cancer (CRC), we developed an Acoustic Resolution Photoacoustic Microscopy (AR-PAM) system.
Unlike previous studies, our approach can differentiate between benign and malignant regions using only PAM, without requiring assistance from other modalities.
The design of our PAM system is shown in~\cref{fig:system}. The system was used to image healthy, cancerous, and polyp tissues \emph{ex-vivo}. With the help of the system, we were able to identify cancerous regions more accurately. We selected photoacoustic signals from the different tissues, and categorized them as cancerous or normal signals for a binary classification task. We evaluated the performance of various machine learning models on the dataset and conducted both quantitative and qualitative experiments to validate our results.

The system used in this study comprised a 532 nm wavelength laser (CNI Co., Ltd., China), a function generator (Tektronix Technology Co., Ltd., USA), a focused ultrasound transducer (Guangzhou Doppler Electronic Technology Co., Ltd., China), a low-noise amplifier (Photosound Technology Co., Ltd., USA), and a data acquisition card (ADLINK Technology Co., Ltd., China). To enable mechanical raster scanning, a 2D stepper motor (Jiangxi Liansheng Technology Co., Ltd., China) was employed, with its movements synchronized with the data acquisition. The stepper motor and data acquisition were controlled using LabVIEW software (National Instrument Co., Ltd., USA).

The pulse energy of the laser we used was 250 \textmu{}J. The pulse width was 6 ns. The beam was coupled into the multimode fiber (MMF) (Daheng Photoelectric Technology Co., Ltd., China). A 10$\times$ objective lens is used to focus the beam into the MMF.
The laser pulse was emitted from the MMF and passed through the focusing lens (Daheng Photoelectric Technology Co., Ltd., China.). The energy of the laser pulse irradiated onto the sample was 42 \textmu{}J.
According to our calculation, the unit energy density is lower than the MPE standard of ANSI (20 mJ/cm$^{2}$). The Function Generator is used to generate a pulse signal at 1 kHz to trigger the laser and data acquisition card (DAQ).

The system utilized a two-dimensional stepping motor to enable mechanical raster scanning. A focused ultrasonic transducer, which was coaxial with the beam and presented a reflection acceptance mode, was used to detect the excited photoacoustic (PA) signal. To compensate for the relatively weak PA signal, a low-noise amplifier (Photosound Technologies, Inc., USA) was integrated into the PAM system. The amplifier was capable of amplifying the PA signal by approximately 40 dB with low noise. Finally, the PA signals were recorded using the DAQ.

\begin{table*}[h]
    \centering
    \caption{Model comparison with quantitative metrics evaulated on eval dataset for out-of-distribution analysis. AUROC, Area Under Receiver Operating Characteristic Curve; AP, Average Precision; PPV, Positive Predictive Value; NPV, Negative Predictive Value; LR, Logistic Regression; $k$-NN, $k$ Nearest Neighbours; MLP, Multi-Layer Perceptron; LDA, Linear Discriminant Analysis; QDA, Quadratic Discriminant Analysis. Models are sorted by a decreasing order of accuracy. The Best values are marked with bold symbols.}
    \label{tab:eval-metrics}
    \begin{tabular}{lccccccccc}
        \hline
        \multicolumn{1}{c}{Classifier} & Accuracy & AUROC & AP & Sensitivity & Specificity & PPV & NPV & F1 Score & Brier Score \\ \hline
        XGBoost & \textbf{0.813766} & 0.773565 & 0.852888 & 0.909084 & 0.545104 & 0.849234 & 0.680223 & \textbf{0.878140} & 0.142804 \\
        LightGBM & 0.809212 & \textbf{0.798137} & \textbf{0.890752} & 0.915388 & 0.509943 & 0.840381 & \textbf{0.681350} & 0.876282 & \textbf{0.139946} \\
        AdaBoost & 0.807548 & 0.763058 & 0.835680 & 0.904362 & 0.534669 & 0.845629 & 0.664818 & 0.874010 & 0.238421 \\
        Decision Tree & 0.752896 & 0.681107 & 0.819002 & 0.828200 & 0.540643 & 0.835575 & 0.527520 & 0.831872 & 0.248738 \\
        RBF SVM & 0.738124 & 0.500000 & 0.738124 & \textbf{1.000000} & 0.000000 & 0.738124 & NaN & 0.849334 & 0.249850 \\
        CatBoost & 0.734679 & 0.629709 & 0.779491 & 0.926843 & 0.193043 & 0.764003 & 0.483523 & 0.837582 & 0.183587 \\
        Random Forest & 0.729075 & 0.757911 & 0.887923 & 0.783024 & 0.577013 & 0.839170 & 0.485463 & 0.810125 & 0.189842 \\
        MLP & 0.689590 & 0.696637 & 0.828363 & 0.735460 & 0.560302 & 0.825008 & 0.429043 & 0.777664 & 0.300092 \\
        Naive Bayes & 0.658264 & 0.716988 & 0.855240 & 0.665120 & 0.638941 & 0.838508 & 0.403669 & 0.741817 & 0.301963 \\
        LR & 0.654997 & 0.673890 & 0.784996 & 0.667883 & 0.618677 & 0.831558 & 0.397918 & 0.740787 & 0.280697 \\
        Linear SVM & 0.641393 & 0.694215 & 0.793538 & 0.631184 & 0.670170 & 0.843600 & 0.391977 & 0.722094 & 0.282323 \\
        LDA & 0.618760 & 0.748777 & 0.825527 & 0.573425 & \textbf{0.746541} & \textbf{0.864440} & 0.383060 & 0.689483 & 0.319179 \\
        QDA & 0.543890 & 0.539292 & 0.773327 & 0.587992 & 0.419584 & 0.740623 & 0.265413 & 0.655541 & 0.452583 \\
        $k$-NN & 0.489139 & 0.465919 & 0.727621 & 0.497130 & 0.466616 & 0.724292 & 0.247672 & 0.589587 & 0.447229 \\ \hline
    \end{tabular}
\end{table*}

The colorectal samples used in this study were obtained from patients treated at the First Medical Centre of the Chinese PLA General Hospital in Beijing. A total of 22 patients were included in the study, and a total of 64 excisions were carefully classified into three categories: healthy, border, and cancerous tissues, by medical professionals. This study was approved by the ethics committee of the hospital, and all patients provided informed consent by signing consent forms.

We utilized our proposed PAM system to scan the colorectal tissues. For each colorectal tissue, we firstly selected a region of interest (ROI), and the field of vision (FOV) is $10\times 10$ mm$^2$. Scans of 100 rows, 250 columns were performed inside these RoIs respectively, and the step sizes were 100 \textmu{}m and 40 \textmu{}m respectively. The center frequency of the focused ultrasonic transducer is 10 MHz. The fractional bandwidth is about 60\%, and the focal length is 18 mm. The length of each signal produced by the PAM system was 1000 with sampling frequency at 80 MHz.

We used the MAP algorithm to determine the value of the pixels within these PAM images.
Suppose that the scans were perform $H$ rows $W$ columns. The size of the deserved images should be $H\times W$. Each signal located in row $i$ column $j$, denoted by $s_{ij} (t)$, was first processed with Hilbert transform~\cite{Oppenheim1999DiscretetimeSP},
\begin{equation}
    \hat{s}_{ij} (t) = \mathcal{H} \left[ s_{ij} (t) \right] = \frac{1}{\pi} \int_{-\infty}^{+\infty} \frac{s(\tau)}{t - \tau} \mathrm{d} \tau,
\end{equation}
where $i = 1, 2, \ldots, H$, $j = 1, 2, \ldots, W$. The exact value of each pixel in the image $I$ was determined by the maximum amplitude of the Hilbert transform,
\begin{equation}
    I_{ij} = \max_t \left| \hat{s}_{ij} (t) \right|.
\end{equation}
By calculating the MAP algorithm, we can obtain the PAM result of a biological tissue. The implementation of the MAP algorithm and visualizations of the human tissues were performed on MATLAB software (MathWorks Co., Ltd., USA).

Photographs and MAP imaging results of the tissues are presented in~\cref{fig:image-results}. The images demonstrate that regions with healthy tissues have strong PA signals, indicating strong optical absorption at 532 nm wavelength, while signals from cancerous areas generate relatively weak signals. The MAP results of boundary and cancer areas clearly separated, making it easy to distinguish between cancerous and healthy areas. The cancerous areas appear as whitish tissues, while the healthy areas appear as hematic red tissues in the photographs.

We compared the performance of 14 different machine learning models using a dataset of 118,687 signals selected from 175,000 signals from 7 colorectal tissue samples. The signals were divided into two classes: cancerous and normal tissues. Out of all the signals, 56,387 signals were from cancerous areas, while 62,300 signals were from healthy regions. To train and test the models, we used 75\% of the dataset as the training data and the remaining 25\% as the test dataset for an in-distribution evaluation.
In order to further evaluate the generability of the classifiers, we extracted 50,501 signals out of 5 samples, which were samples different from the former dataset, as an out-of-distribution evaluation. The evaluation dataset consisted of 37276 cancer signals and 13225 normal signals.

\begin{algorithm}[b]
    \caption{Data pipeline for preprocessing}\label{alg:data-pipeline}
    \begin{algorithmic}[1]
    \Procedure{Preprocessing}{$s$}
    \State $s \gets \frac{s - \text{mean}(s)}{\text{std}(s)}$ \Comment{Standard scaler}
    \State $\hat{s} \gets \gH (s)$ \Comment{Hilbert transform}
    \State $y \gets  \gF (\hat{s}) $ \Comment{Fourier transform}
    \State \textbf{return} $\left| y \right|$ \Comment{Return the amplitude of complex signal}
    \EndProcedure
    \end{algorithmic}
\end{algorithm}

\begin{figure*}[t]
    \centering
    \subfloat[ROC curve on test dataset]{
    \includegraphics[width=0.32\textwidth]{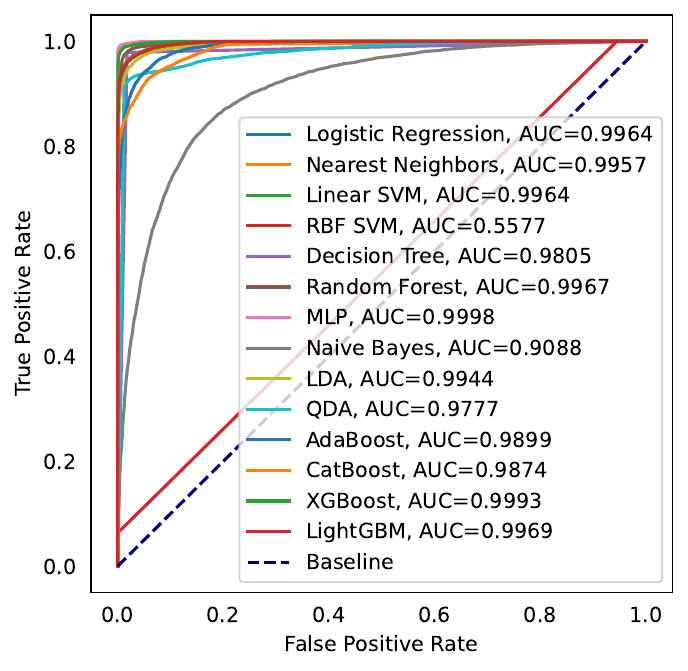}\label{fig:roc}
    }
    \subfloat[PRC curve on test dataset]{
    \includegraphics[width=0.32\textwidth]{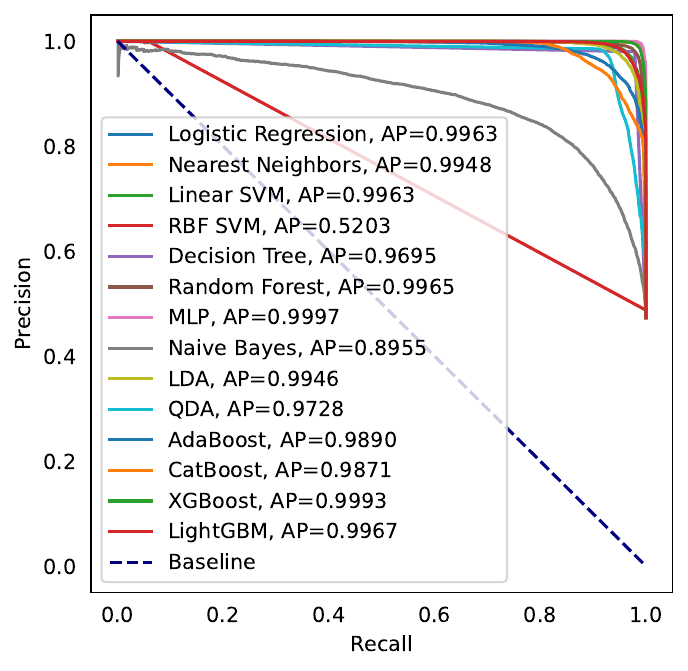}\label{fig:prc}
    }
    \subfloat[ROC curve on evaluation dataset]{
    \includegraphics[width=0.32\textwidth]{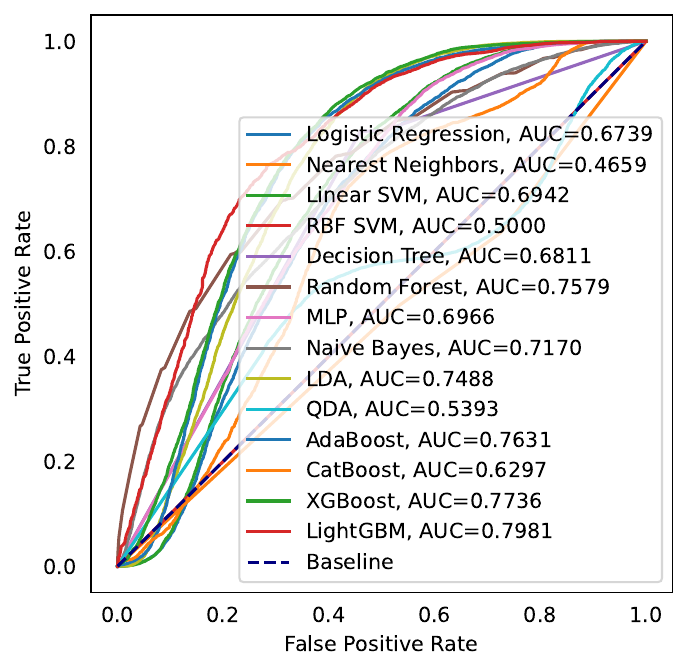}\label{fig:roc_eval}
    }
    \caption{Model comparison on the benign and malignant binary classification problem. ROC, Receiver Operating Characteristic curve; PRC, Precision Recall Curve.}\label{fig:clf-comp}
\end{figure*}

A variety of machine learning classifiers were involved in this experiment, including Logistic regression (LR); $k$ Nearest Neighbors ($k$-NN) with $k = 2$;
Support Vector Machine (SVM)~\cite{Platt1999ProbabilisticOF}, including linear SVM and SVM with Gaussian radial basis function kernel (RBF SVM); Decision Tree~\cite{Loh2011ClassificationAR}; Random Forest~\cite{Breiman2001RandomF}; Multi-Layer Perception (MLP); Na\"ive Bayes; discriminant analysis methods including Linear Discriminant Analysis (LDA)~\cite{fisher1936use} and Quadratic Discriminant Analysis (QDA); and gradient boosting algorithms, including Adaboost~\cite{Freund1997ADG}, CatBoost~\cite{Dorogush2018CatBoostGB}, XGBoost~\cite{Chen2016XGBoostAS} and LightGBM~\cite{Ke2017LightGBMAH}.

We stochastically split the training dataset and test dataset before training, and all the machine learning models were trained and tested on the same data.
We constructed a data preprocessing pipeline for all the data, as is shown in~\cref{alg:data-pipeline}.
\cref{fig:clf-comp} displays the receiver operating characteristic curve (ROC) and the precision recall curve (PRC) on test dataset and the ROC on evaluation dataset. The area under the ROC curve (AUROC) is a widely used metric for evaluating classifiers. In~\cref{fig:roc}, we observe that all 14 machine learning models achieved impressive results on the binary classification task. RBF SVM had the lowest AUROC of 0.5577, while XGBoost had the highest AUROC of 0.9993 among all the models. Similarly, the area under the PRC, also known as average precision (AP), is another popular metric for assessing classifiers. XGBoost again had the best AP score.

However, all the classifiers didn't perform that well on the evaluation dataset.
Based on~\cref{tab:eval-metrics} and~\cref{fig:roc_eval}, we can see that the XGBoost and LightGBM classifiers had the highest accuracy among all the models tested. XGBoost also had the highest F1 Score, which is a metric that combines both precision and recall. LightGBM had the highest AUROC and AP, which are metrics that evaluate the model's ability to discriminate between positive and negative samples. However, XGBoost had a slightly better PPV (positive predictive value) than LightGBM, which means that XGBoost was better at correctly predicting positive samples. On the other hand, the $k$-NN classifier had the lowest accuracy and AUROC among all the models tested. RBF SVM had the highest sensitivity of 1.0, meaning it correctly identified all positive samples, but at the cost of low specificity and NPV scores. The reason for this was that RBF SVM predicted all the signals to be cancer signals on the evaluation dataset, indicating low generalization capability and robustness.


In this study, we developed an AR-PAM system for the detection of CRC. Our system demonstrated high contrast imaging capabilities as evidenced by the clear boundary between cancerous and healthy tissues shown in the \emph{ex vivo} MAP results. To classify benign and malignant signals, we created a dataset and evaluated the performance of several machine learning models. Our results indicate that these models can accurately distinguish signals from benign and malignant regions, suggesting their potential use in CRC diagnosis. Moving forward, we plan to further improve the system's imaging capabilities in 3D through additional studies.


\begin{backmatter}

\bmsection{Funding} This research was funded by National Natural Science Foundation of China (61805139), United Imaging Intelligence (2019X0203-501-02), and Shanghai Clinical Research and Trial Center (2022A0305-418-02).

\bmsection{Acknowledgments} We are grateful to the Chinese PLA General Hospital for cooperation.

\bmsection{Disclosures} The authors declare no conflicts of interest.


\end{backmatter}

\bibliography{reference}

\bibliographyfullrefs{reference.bib}

\ifthenelse{\equal{\journalref}{aop}}{%
\section*{Author Biographies}
\begingroup
\setlength\intextsep{0pt}
\begin{minipage}[t][6.3cm][t]{1.0\textwidth} 
  \begin{wrapfigure}{L}{0.25\textwidth}
    \includegraphics[width=0.25\textwidth]{john_smith.eps}
  \end{wrapfigure}
  \noindent
  {\bfseries John Smith} received his BSc (Mathematics) in 2000 from The University of Maryland. His research interests include lasers and optics.
\end{minipage}
\begin{minipage}{1.0\textwidth}
  \begin{wrapfigure}{L}{0.25\textwidth}
    \includegraphics[width=0.25\textwidth]{alice_smith.eps}
  \end{wrapfigure}
  \noindent
  {\bfseries Alice Smith} also received her BSc (Mathematics) in 2000 from The University of Maryland. Her research interests also include lasers and optics.
\end{minipage}
\endgroup
}{}

\end{document}